\documentclass{ieeeaccess}
\usepackage{cite}
\usepackage{amsmath,amssymb,amsfonts}
\usepackage{algorithmic}
\usepackage{graphicx}
\usepackage{textcomp}

\usepackage{mathtools}
\DeclareUnicodeCharacter{2212}{-}
\usepackage{booktabs}

\DeclarePairedDelimiterX{\infdivx}[2]{(}{)}{%
  #1\;\delimsize\|\;#2%
}

\def\BibTeX{{\rm B\kern-.05em{\sc i\kern-.025em b}\kern-.08em
    T\kern-.1667em\lower.7ex\hbox{E}\kern-.125emX}}
\begin{document}
\history{}
\doi{}

\title{Emotion Analysis using Multi-Layered Networks for Graphical Representation of Tweets}
\author{\uppercase{Anna Nguyen}\authorrefmark{1},
\uppercase{Antonio Longa}\authorrefmark{2,3}, \uppercase{Massimiliano Luca }\authorrefmark{1,2,4}, \uppercase{Joe Kaul}\authorrefmark{1} \uppercase{and} \uppercase{Gabriel Lopez}\authorrefmark{1}}
\address[1]{Pulse.io, London SE1 2SA, United Kingdom}
\address[2]{Fondazione Bruno Kessler, Trento, Italy }
\address[3]{Unversity of Trento (Italy)}
\address[4]{Free University of Bolzano, Bolzano, Italy}

\markboth
{Nguyen \headeretal: Emotion Analysis using Multi-Layered Networks for Graphical Representation of Tweets}
{Nguyen \headeretal: Emotion Analysis using Multi-Layered Networks for Graphical Representation of Tweets}
\begin{abstract}
Anticipating audience reaction towards a certain piece of text is integral to several facets of society ranging from politics, research, and commercial industries. Sentiment analysis (SA) is a useful natural language processing (NLP) technique that utilizes both lexical/statistical and deep learning methods to determine whether different sized texts exhibit a positive, negative, or neutral emotion. However, there is currently a lack of tools that can be used to analyze groups of independent texts and extract the primary emotion from the whole set. Therefore, the current paper proposes a novel algorithm referred to as the Multi-Layered Tweet Analyzer (MLTA) that graphically models social media text using multi-layered networks (MLNs) in order to better encode relationships across independent sets of tweets. Graph structures are capable of capturing meaningful relationships in complex ecosystems compared to other representation methods. State of the art Graph Neural Networks (GNNs) are used to extract information from the Tweet-MLN and make predictions based on the extracted graph features. Results show that not only does the MLTA predict from a larger set of possible emotions, delivering a more accurate sentiment compared to the standard positive, negative or neutral, it also allows for accurate group-level predictions of Twitter data.
\end{abstract}


\maketitle
\section{Introduction}
Social media has become an integral part of society. Particularly, media outlets such as Twitter, are often used for spreading information, advertising, and curating communities. 
Researchers have used Twitter data to make predictions on many scenarios, for instance, election results~\cite{coletto2015electoral}, analyze product opinions~\cite{6726818}, forecast box office revenue~\cite{5616710},  measure public response to COVID-19~\cite{xue2020twitter}, and predict how people move in cities \cite{luca2021survey}. 

Another task that commonly leverage{s} Twitter data is sentiment analysis (SA) \cite{birjali2021comprehensive}. SA is a natural language processing (NLP) technique often used to determine whether a text falls into one of three categories: positive, negative, or neutral~\cite{mejova2009sentiment}. As those three sentiments can be quite unspecific, it is important to find ways of extracting a wider range of emotions from text. Additionally, while there have been algorithms developed to accommodate different sized texts, there is limited research on models which extract the overall sentiment of a collective group of independent texts. The current research, seeks to have a deeper understanding of the overarching sentiment of a group of unrelated texts as it adds to the real-life applicability of sentiment analysis and allows for a whole new understanding of attitudes. This practice can be quite useful for industry professionals as it alleviates the clients' responsibility to interpret results or manually adjust the output of a traditional SA model which is limited to predictions on individual tweets. 


{Current methodologies, associate to a set of texts discussing the same topic the average predicted emotion. For example, given ten tweets discussing a topic $\mathcal{T}$, with six positive tweets and four negative tweets, will be considered positive.} Such methodologies neglect the impact of potential associations that come out of independent entities that are placed together. Realistically, individuals looking at a group of items will formulate associations between them, even if they do not exist. Similarly, independent items placed next to each other will not be evaluated without bias, but rather an opinion, attitudes, and behaviors are affected by the surrounding environmental information{\cite{cassidy2013environmental}}. 
Psychological studies support this claim that individuals can be influenced by the proximity effect, where purchasing behaviours are dependent on an items' relative website positioning \cite{breugelmans2007shelf}. Therefore, the process of analyzing tweets individually and then aggregating the results is not conducive to modeling the correct behavior of the general public. Which is the behavior that is most useful for industries trying to understand the sentiment of their clientele. Take as {another} practical example, if the frequency distribution of labels is 51\% negative and 49\% positive { for a different topic $\mathcal{T}_2$}. Presenting this ratio to a user who has requested the general emotion from a group of tweets is not useful, because the labels are almost evenly distributed, and it does not help the client to understand what the broader group-level sentiment is. Subsequently, if the label were to be returned as negative, that can also be misleading as the emotion intensity of the text is not examined. For example, the 49\% of happy tweets could be extremely positive, whereas a number of the negative tweets could be leaning toward neutral but contain enough negativity to become classified as negative. As stated before, since individuals have a tendency to evaluate using information from the whole set of tweets, this distinction between the emotion intensity can lead a client to believe that the overarching emotion is positive. 

As these nuances are difficult to detect and adjust for, it is important to have models capable of formulating a deeper understanding of the relationship between a group of independent tweets. Evaluating how tweets in a set interact with each other in order to give the impression of a certain emotion is different from separately classifying them and then aggregating the results. 




In the last decades, graph analysis has been involved in many domains such as: molecular biology\cite{huber2007graphs}, face-to-face interactions \cite{longa2022efficient,longa2022neighbourhood}, for social networks \cite{barnes1969graph,brandes2013social}, {crime \cite{calderoni2020robust, ficara2021criminal}} and many others \cite{luca2021leveraging, cencetti2021digital}. 
For NLP problems, graphs are very useful for representing text because the syntactic relationship between words in a sentence can naturally form a graph structure \cite{nastase2015survey}. Furthermore, multi-layer networks (MLN) which are formed by connecting several systems together, can be used to represent inter-dependent aspects of a heterogeneous group of tweets \cite{singh2020sentiment}.

For that reason, the proposed methodology represents a group of unrelated tweets using a multi-layer graph network. The graph structure can help to emulate a number of different tweets of a similar sentiment, as a single entity and makes it possible to understand the underlying associations that lead to a group of tweets sharing an overarching emotion. Additionally, the Tweet-MLNs will resolve many of the practical issues with using social media text, as they will be capable of capturing meaningful relationships from colloquial language and unseen tweet assets such as hashtags, emojis, or mentions because of how the data is structured inside the graph. Furthermore, due to the multi-dimensionality of the MLN different relationships between different features of a tweet can be examined at the same time. For example, the relationship among hashtags that occur within the same tweet, and are repeated in other tweets can be modeled simultaneously with the syntactic relationship between keywords. 


In summation, this work uses a MLN graph representation of Twitter data and is passed through a GNN to make group-level predictions of the overall emotion of a group of independent tweets. Additionally, the dataset includes a wider range of emotions: Happy, Angry, Bad, Fearful, Sad, and Surprised. 

The rest of the paper is structured as followed. {In Section \ref{sc:rel_work}, we shortly introduce some related work related to sentiment analysis. In section \ref{sec:rw} we introduce useful concepts and notations about graphs and graph neural networks.} Section \ref{sec:implementation} breaks down the steps toward executing the proposed model and also provides an overview of the network architecture. Section \ref{sec:et} details experimental tests used to evaluate the network, while the results of these tests are analyzed in section \ref{sec:r}. Finally, the paper is concluded in section \ref{sec:c}.

{ \section{Related Work}
\label{sc:rel_work}
The rapid diffusion of ubiquitous technologies like mobile phones also came with the unprecedented opportunity for companies to analyze the opinion, sentiments and perceptions of people with respect to a specific product or topic. Sentiment Analysis (SA) is the task consisting of the extraction of such insights. In recent years, several applications have seen success in their ability to perform such classifications to a high degree of accuracy. There are two main branches of SA, one using lexical rules and the other using machine learning techniques (e.g., \cite{medhat2014sentiment,birjali2021comprehensive}). Lexical methods study the semantic information already present inside a piece of text and use statistics to extract features (e.g., \cite{taboada2011lexicon}). Machine learning methods are framed as traditional text classification problems (e.g., \cite{le2015twitter}). Usually, ML-based method leverage datasets \(D = \{(X_1, Y_1), ...,(X_n, Y_n)\}\) where X is a textual resource and Y is a class (positive, negative, or neutral). Then a classification model is trained on the data, and used to predict labels for unseen pieces of text. Hybrid methods combine the two techniques (e.g., \cite{zainuddin2018hybrid}). While there are many deep learning approaches to classify the sentiment of texts \cite{habimana2020sentiment}, in this paper, we focus on the role of transformers.
Recently, transformer networks \cite{vaswani2017attention} reached promising results and are widely adopted to solve NLP and Computer Vision tasks \cite{lin2021survey, khan2021transformers}. 
BERT \cite{devlin2018bert} is one of the first transformers aiming to solve Natural Language Understanding problems. It relies on a set of encoders and self-attention mechanisms. While it was pretrained to be a language model and to perform next sentence prediction \cite{devlin2018bert}, it has been widely adopted to solve SA problems achieving significantly high performances \cite{xu2019bert, sun2019utilizing, hoang2019aspect, li2019exploiting, pota2020effective}. 

In this paper, we use BERTSent \cite{govindarajan2020help} and XLM-roBERTa \cite{conneau2019unsupervised} as baselines for standard SA and to support the group-level prediction we are proposing. 
}
\section{Background}\label{sec:rw}

\subsection{Multi-Layered Networks}
Any graph \(G\) consists of vertices and edges where \(G=(V, E)\) and where \(V = {v_1,...,v_N}\) is connected by edges \(E = {e_,...,e_{|N|}}\). A graph is weighted if there exists a function that assigns weights to edges connecting two nodes in the set of edges ($(u,v) \in E$). Otherwise it is assumed that a graph is unweighted. 
A graph is undirected if there is a symmetric relationship between nodes and is direct if there is a no symmetry between nodes. Additionally, in an undirected graph, each node points exclusively to another node. 
Graphs are homogeneous if nodes and edges point to one type of entity, and heterogeneous if it contains varying types of nodes and edges. 

Multi-layered Networks are a collection of separate graph structures, or "layers". Each layer shares a connection to at least one other layer in the network. In this way, it is possible to illustrate the relationship between independent graphs that are somehow related. Many real-world scenarios are heterogeneous systems that interact with each other rather than one holistic network, which is why MLNs have become increasingly popular in graphical research \cite{boccaletti2014structure}.  Since the individual layers can capture nuances of its' own environment but the connections across layers still contain information about the relationship between graph entities, the MLN is a practical way of modeling disjointed information of more realistic and complex problems such as social ties \cite{brodka2012multi}, biological systems \cite{lee2020heterogeneous}, and certain inter-dependent infrastructures \cite{chen2016fascinate}. 

Specifically for Twitter data, choosing to reconstruct text into a multi-layered graphical representation can provide many benefits. The multi-layered structure of the graph allows for isolation of tweet-features that are not specifically related to the syntax, such as hashtags. These features can be modeled in their own network representation, but remain connected to the separate graph which models the syntax and key-words in a tweet. Additionally,  \cite{singh2020sentiment} found that the graph structure of a tweet can help to improve classification and handle noisy, unspecific and linguistic challenges compared to other tweet-prediction models. 

One can define the MLN as a set of {$M$} networks or graphs each depicting a separate interactions within their own graph structure but containing some level of cross-layer dependency or association between nodes in a separate graph layer \cite{medhi2000multi}. Three primary types of MLNs include: 
\begin{enumerate}
\item Multiplex 
    \begin{enumerate}
        \item Multiplex Networks creates edges only between repeating nodes in different layers. 
    \end{enumerate}
\item Multi-slice Network 
    \begin{enumerate}
        \item Multi-slice Networks are temporal networks, where edges are present at specific times \(t\). 
    \end{enumerate}
\item Networks of Networks 
    \begin{enumerate}
        \item In a Network of Networks each layer is formed by a different set of nodes and at the layer level there is network itself interacting with other networks in other layers.
    \end{enumerate}
\end{enumerate}

The current research adapts the Network of Networks type MLN with 3 layers each formed of varying sets of nodes from the same group of tweets. Then a GNN is applied at each layer level whose output interacts with the other layers and eventually returns a single prediction for the group of tweets using information from all 3 layers. 

\subsection{Graph Neural Networks}
Compared to other data types such as images, which have a square and fixed structure that can be easily passed through traditional convolutional neural networks (CNN), graph data is more irregular. For example, a graph can contain any number of unordered nodes, and nodes from separate graphs do not have a standardized number of neighbors, making convolutions much more difficult compared to image data which has a fixed neighborhood. Furthermore, instances of nodes are dependent on each other through links that form edges in the graph. 

The deep learning model GNN is applied to solve various graph related tasks and handle the issues with irregularity described above. Without having to change the topology of a graph input, the GNN can perform transformations on all graph attributes while maintaining graph connectivity and symmetry \cite{scarselli2008graph}. In order to do so, GNNs start by converting nodes to recurrent units. Additionally, edges are translated into feed forward neural networks. Message passing between all nodes allow for the extraction of different embedding vectors. Then these embedding vectors are aggregated to formulate a single vector. The current paper will apply a GNN to a graph-level classification task. Graph-level classification is a graph task which attempts to predict the label of an entire graph rather than individual nodes. GNNs have shown improved accuracy on graph classification tasks \cite{errica2019fair} and more relevantly, classification of text-graphs \cite{huang2019text}. A number of different variants of the GNN and different graph convolution layers have been developed overtime and can be applied to graph-level classification. { This study focuses on three types of GNN which are explained below. }

\subsubsection{GCNConv}
Graph convolutional networks (GCNs) are computationally efficient and have been shown to extract powerful node representations \cite{kipf2016semi}. Using a similar idea to traditional convolutions used in CNNs for images, but adjusting from the irregular structure of a graph, the GCN extracts information from a certain nodes' neighborhood and then uses an aggregation function which combines that neighborhood information to be put into a neural network. This process is repeated for each node in the graph. 

The propagation rule applied in the current paper using the GCNConv is as followed: 

\begin{equation}
\mathbf{X}^{\prime} = \mathbf{\hat{D}}^{-1/2} \mathbf{\hat{A}}
        \mathbf{\hat{D}}^{-1/2} \mathbf{X} \mathbf{\Theta}
\end{equation}

{where \(\mathbf{X}^{\prime}\) is the convoluted signal matrix,} \(\mathbf{\hat{A}} = \mathbf{A} + \mathbf{I}\) is the adjacency matrix and identity matrix for the current layer in the MLN. Summing the adjacency matrix and identity matrix creates a self-loop that includes the current node into the aggregation function. 
\(\hat{D}_{ii} = \sum_{j=0} \hat{A}_{ij}\) is the corresponding diagonal matrix. {\(\mathbf{\Theta}\) is a matrix of filter parameters.}

\subsubsection{GATv2Conv}
Graph Attention Networks (GAT) have convolution layers that place different weights on various nodes features, thus paying “attention” to only the most important features in a specific graph \cite{ velivckovic2017graph} and help us to capture a more accurate representation of graph nodes across the different layers. The GATv2Conv is an improvement on the previous model and fixes the static attention problem present in traditional GAT networks \cite{brody2021attentive}. 

GATv2Conv layers were used to extract updated node features \(x_i\), with shape: \((|\mathcal{V}_i|, H * F_{out})\) where \(H\) is number of multi-head attentions and \(i\) is the current layer. This new shape is fed into each subsequent GATv2Conv layer. The current work used 5 heads. 

Since the tweet-graphs did not contain edge features the following formula was used to calculate attention coefficients: 

\begin{equation}\label{formula:gat}
\alpha_{i,j} =
        \frac{
        \exp\left(\mathbf{a}^{\top}\mathrm{LeakyReLU}\left(\mathbf{\Theta}
        [\mathbf{x}_i \, \Vert \, \mathbf{x}_j]
        \right)\right)}
        {\sum_{k \in \mathcal{N}(i) \cup \{ i \}}
        \exp\left(\mathbf{a}^{\top}\mathrm{LeakyReLU}\left(\mathbf{\Theta}
        [\mathbf{x}_i \, \Vert \, \mathbf{x}_k]
        \right)\right)}
\end{equation}

{where a is the attention score \(\mathbf{\Theta}\) is a matrix of filter parameters, \(x_i\) is the input feature matrix and \(x_j\) is the output feature matrix.}
\subsubsection{GraphConv}
GraphConv uses information from higher-order graph structures at various different thresholds \cite{morris2019weisfeiler} for graph tasks. The researchers found that this information is useful in both regression and graph classification problems. The formula for the GraphConv operator is as followed: 

\begin{equation}
\mathbf{x}^{\prime}_i = \mathbf{W}_1 \mathbf{x}_i + \mathbf{W}_2
        \sum_{j \in \mathcal{N}(i)} e_{j,i} \cdot \mathbf{x}_j    
\end{equation}

where the edge weight \(e_{j,i}\) goes from the original node \(j\) to
target node \(i\).

\section{Implementation of Tweet-MLN for Group-Level Emotion Analysis} \label{sec:implementation}
The following section provides a detailed breakdown of the proposed Multi-Layered Tweet Analyser (MLTA) used for social media text-classification. It begins by describing how tweets were transformed into their respective multi-layer network representation or Tweet-MLN. Then it maps out embedding techniques used for nodes inside the graph. Finally, the overall architecture of neural networks used for training is detailed. 

\subsection{Constructing Multi-Layered Networks Using Tweets}
\Figure[t!](topskip=0pt, botskip=0pt, midskip=0pt){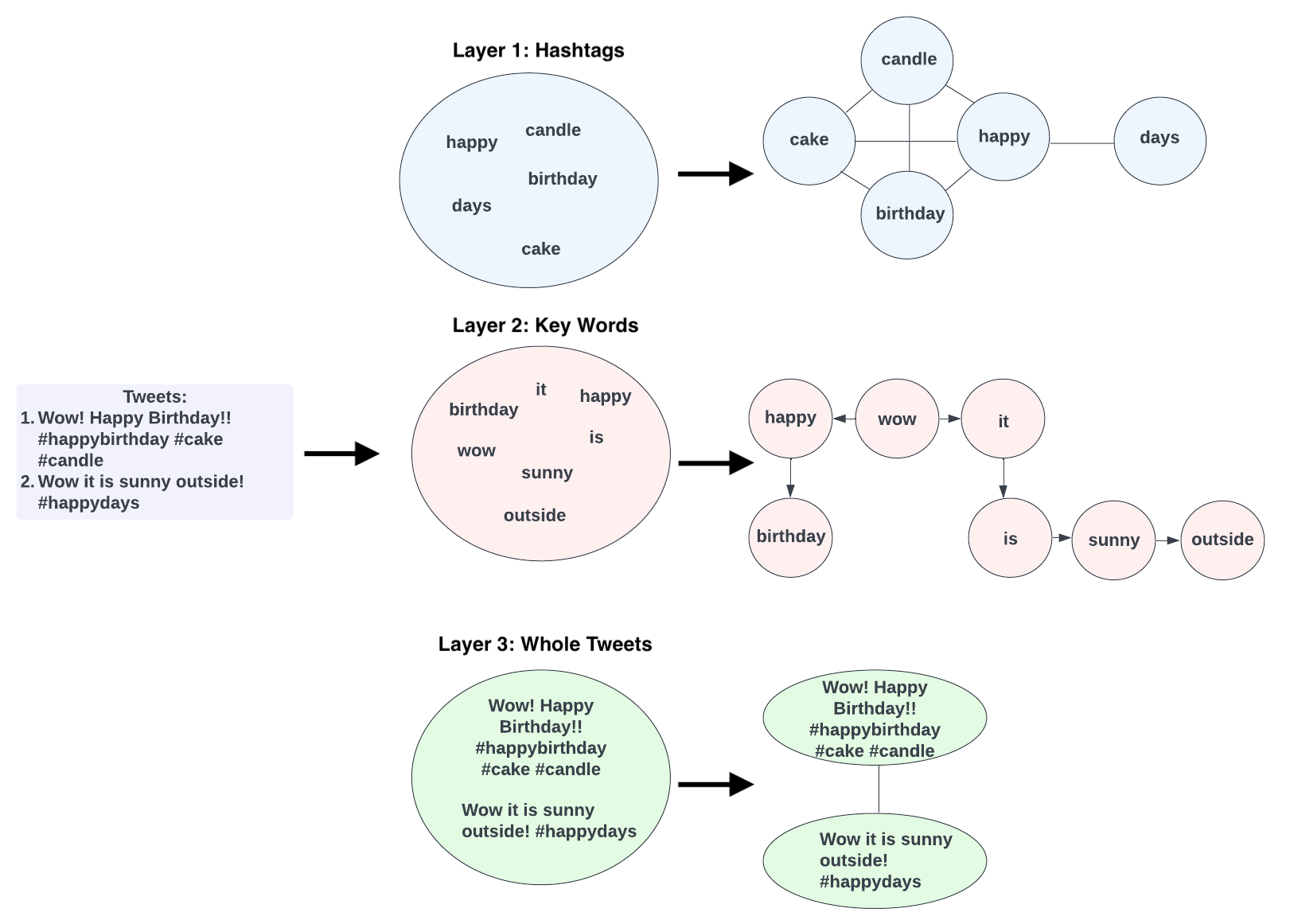}
{Tweet Multi-Layered Graph Structure. On the left is a set of tweets. Each tweet is pre-processed and subsections of the tweet are separated out into different groups for each layer of the MLN. Layer 1 receives all the hashtags present in the set of tweets. Layer 2 receives all key words with hashtags removed. Layer 3 receives the whole set of tweets kept in their original format. After the tweets are organized into their respective groups, graphs are constructed from the group entities. The Layers 1 \& 3 are undirected graphs and Layer 2 is a directed graph. \label{fig:MLN}}

In order to construct the Tweet-MLN, twitter sets are generated which consist of 300 tweets that share the same class label. This number seemed reasonable for practical use of a company in the future, as clients are more likely to analyze large bundles of tweets, if they are searching for an overall general emotion. However, it is not too large that a smaller client would not be able to submit a sufficient number of tweets. Further breakdown of the dataset can be found below in section \ref{sub-dataset}. The following information is extracted from every tweet within a set: 
\begin{itemize}
    \item Hashtags 
    \item Key-words (individual words that make up full body text)
    \item Full-body text 

\end{itemize}

These components are used to formulate nodes in different layers of the Tweet-MLN. Therefore, one cluster of 300 tweets will result in a single Tweet-MLN which holds the above information from all of the individual tweets within that set. The class label that is shared across each individual tweet, becomes the overall class evaluated by the network.  

Following traditional graph classification problems each Tweet-MLN is given a single emotion label, and is held in set \(C=\{c_1,c_2,c_3,…,c_n\}\). 

The first graph layer \(L1\) is an undirected graph that consists of only hashtags. Nodes are individual hashtags and edges are created between hashtag nodes if they appear in the same tweet or if they appear in a tweet that shares a hashtag. 
{Hashtags that contain multiple words such as \emph{\#happybirthday} 
are broken down into separate word tokens \emph{happy} and \emph{birthday} for two reasons. 
The first being that it enables us to extract more meaningful and accurate word embeddings as individual words in their original format are more likely to be found in a pre-trained embedding corpus. The second being that this graphical architecture allows for the study of a complex relationship between specific words. 
While the hashtags \#happybirthday and \#happydays are intended to convey different expressions, the word happy still carries the same practical meaning in both. 
Therefore, this structure allows us to see how other words used around the word happy may be related and how a word with one meaning can be intermingled with others.While this may lead to repeated nodes within the current layer and the next, no physical edge information between the two layers is presented to the network, therefore, making it still a Network of Networks MLN.}

The second layer \(L2\) is a directed graph that contains all individual words inside the tweet excluding the hashtags. Each key-word is a node and each directed edge goes from the start of the sentence to the subsequent word until the end of the sentence is reached. If there is a hashtag in the middle of the sentence, then that hashtag is removed and the edge is connected to the word following the hashtag. The third layer \(L3\) is an undirected graph and consists of the entire raw tweet as a node. Edges are placed between tweets that share hashtags. See Figure \ref{fig:MLN} for a visual representation of how tweets were broken down into each layer. { After the Tweet-MLNs were constructed, the text contained inside the graph nodes were converted into word-embeddings, which are detailed in the following subsection \ref{sec:wordemb}}

Let \(M\) represent a single Tweet-MLN object. \(M\) can be defined as \(M= (V,E,\mathcal{L})\) where \(\mathcal{L}\) denotes a set \(\mathcal{L}=\{L1,L2,L3\}\) which holds the three tweet text layers described above. \(V = \{V_1..V_i\}\) where \(V_i\) are the vertices in layer \(i\) and \(E\) denotes the set of edges for each layer and there are no shared vertices between layers. Similar to \cite{singh2020sentiment}, \(M\) includes a mixture of direct and undirected graphs, however rather using a matrix of matrices, the current research keeps each graph layer separated, in order to allow the GNN to extract independent graph features. Then following extraction, features from each graph are concatenated and fed into two fully-connected linear layers which output a prediction based on the conjoined features. 

\subsection{Word Embeddings}\label{sec:wordemb}
Word embeddings also known as word vectors are used to translate words into a usable format for NLP problems. A word embedding is an continuous representation of a word in array format. A neural network can be trained to learn these embeddings so that words with a shared meaning are represented in a similar structure. The current work uses 2 different word-embedding models in order to exploit the different encodings of language in each library. By vertically stacking the word representations, a feature matrix is constructed for each layer of the MLN. 

The first model used is FastText \cite{bojanowski2017enriching} where words are depicted by the summation of character n-grams. Given a corpus with vocabulary of size \(N\) with indexed words \(w \in \{w_1....w_N\}\). FastText aims to maximize the following log-likelihood model: 
\begin{equation}
    \sum_{T=1}^{T} \sum_{c\in{C_t}} log \text{ } p(w_c \mid w_t)
\end{equation}

where \(C_t\) is the context of the word given by the set of indices behind and in front of the \(w_t\). The paramaterization of the word vectors is based on the probability of finding \(w_c\) given the current \(w_t\) \cite{bojanowski2017enriching}. 
The second model used is GloVE \cite{pennington2014glove} which stands for Global Vectors for Word Representation. In this model, the dot product between two word vectors is equal to the log of co-occurrence frequency (or the total number of instances in which two words appear next to each other in a specific context). Therefore given two words inside a corpus of \(w_i...w_j\) words: \[w_i \cdot w_j = log(P(i \mid j))\] 
GloVE takes these co-occurrence probabilities and represents the word embeddings as that ratio and simultaneously captures the global distribution of words \cite{pennington2014glove}.  

Together the two models are able to represent information using different encodings and because they were trained using a different corpus, they allow for us to filter through and make sure that we have a proper word-representation for more of the words in our datasets' vocabulary. We first pass a word through the GloVE model and if the word embedding is not present, it is subsequently sent through to the FastText model. One great component of FastText is that it is able to encode words that were not in the training dataset and match it with similar words from its corpus. This means that out of vocabulary (OOV) ngrams will automatically be mapped to a vector that matches the most similar ngram \cite{bojanowski2017enriching}. 

\subsection{Overall Network Architecture}
\Figure[t!](topskip=0pt, botskip=0pt, midskip=0pt){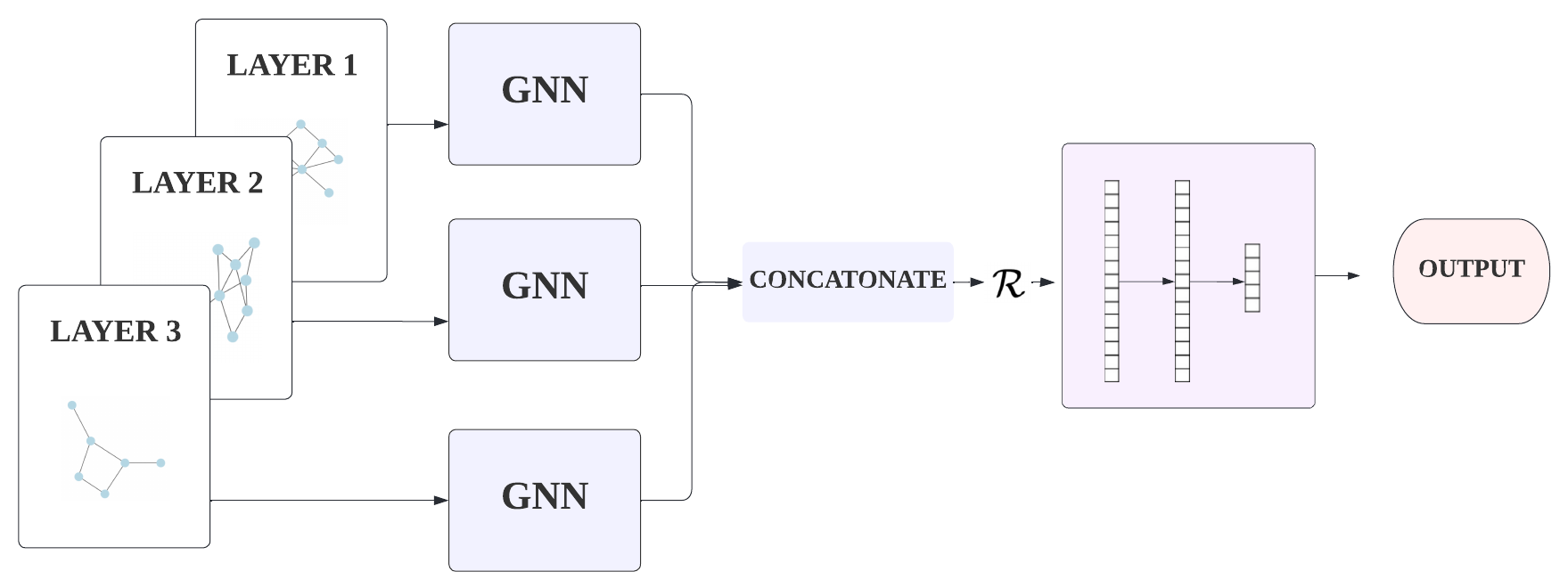}
{Network Architecture of MLTA. Each layer is inputted into its' own GNN before being concatenated and sent into 2 fully-connected linear layers.  \( \mathcal{R}\) is a vector formed by concatenating the extracted features from all 3 layers \label{fig:GATFlow}}

The architecture of the MLTA consists of 3 input layers, one for each layer in the Tweet-MLN, followed by two graph convolution layers, a global mean pooling function, concatenation function, two fully-connected linear layers, and finally one linear output layer. The described architecture of the model is shown in Figure \ref{fig:GATFlow}. 

Three feature matrices containing node embeddings for different layers of the MLN were used as inputs. Each had a shape \((|\mathcal{V}_i|, F_{in})\) where \(|\mathcal{V}_i|\) is number of nodes, \(F_{in}\) is the number of node features, and \(i\) as the subscript representing the corresponding layer.
{As node features are the token-embeddings themselves, the number of node features directly corresponds to the size of the word-embedding vector. Which in this case is 300.}
These inputs were passed through graph convolutions to extract updated node features from the separate entities in the Tweet-MLN. 

An ablation study was conducted on 3 different types of graph convolutions: GCNConv, GATv2Conv, and GraphConv, in order to determine which type of convolution was most suitable for prediction of the Tweet-MLN using a GNN. 

Batched training was used due to the large size of the dataset. Therefore in order to concatenate the extracted features for each layer, we used a global mean pooling function, which outputs a batch-sized value on the graph-level. The function aggregates and averages node features, so that for a single graph \(G_i\) whose output is \(x_{i}\) and shape:  \((|\mathcal{V}_i|, F_{out})\), the mean is computed by:

\begin{equation}
    \mathbf{r}_i = \mathrm{max}_{n=1}^{N_i} \, |x_{n_i}|
\end{equation}

The \(\mathbf{r}_i\) for each of the layers are concatenated to form one input \(\mathcal{R}\): 
\begin{equation}
 \mathcal{R} = \mathbf{r}_i + \mathbf{r}_{i+1}... + \mathbf{r}_n  
\end{equation}

Concatenation of the node features allows the final layers to learn from the joint information of each graph structure. Previous literature supports that feature concatenation has been able to enhance classification accuracy in a number of different machine learning tasks \cite{8351550}, \cite{doukim2021combining}.
\(\mathcal{R}\) is passed through two fully-connected linear layers. Then finally, these node features are given to a linear output layer that returns a single prediction from the 6 emotion labels. 

\subsection{Network Parameters} 
Layers 1, 2, \& 3 were each passed through 2 graph convolutions of starting shape 300 and then got reduced to features of sized 128. After each of the graph convolution layers (for each separate Tweet-MLN layer), a Relu activation function was applied, along with a dropout of 0.5 after the last graph convolution layer. 

Adam optimizer was used and a learning rate of 0.001 was applied. The Tweet-MLN was coded using Pytorch Geometric and NetworkX and the GNN model was constructed using PyTorch.

As the dataset consists of categorical data we calculated the cross entropy loss between the input and target variables using: 

\begin{equation}
\begin{aligned}
          &\ell(x, y) = L = \{l_1,\dots,l_N\}^\top, \\
          &\quad l_n = - w_{y_n} \log \frac{\exp(x_{n,y_n})}{\sum_{c=1}^C\exp(x_{n,c})} 
          &\cdot \mathbb{1}\{y_n \not= \text{ignore\_index}\}
\end{aligned}
\end{equation}

{where \(x\) is the input graph and \(y\) is the target class, \(w\) is the weight, \(N\) is the batch dimension and \(c\) is the total number of classes.}

\section{Experimental Tests}\label{sec:et}
\subsection{Twitter Emotion Dataset}\label{sub-dataset}
Data was obtained from the Twitter API using key words related to the 6 primary emotions found in \cite{ekman1992there} with two alterations: Angry, Bad, Fearful, Happy, Sad and Surprised. Bad was added as a primary emotion to serve as broader category for tweets that did not fit into other negative emotions. Additionally, Fearful and Disgusted were combined into one emotion because there is a large body of research that suggests the two emotions are heavily related \cite{neumann2012priming}. Research has also shown that disgust is is enhanced by fear, but the inverse is not always true \cite{edwards2006experimental}. Which is why, the current research uses Fearful as the primary label.  

The process for data collection is as followed: using the Twitter API, when searching for Angry tweets in addition to the word, angry, the search included others like frustrated, annoyed, betrayed etc. Derivative of the technique used by \cite{plutchik1980general} in which primary emotions are expanded into secondary and tertiary descriptors using a visual wheel representation, we collected tweets commonly associated with each target label based on a "Wheel of Emotions" that has previously been used to aid emotional expression in therapy \footnote{F. Dhuka, “How to Use ‘Wheel of Emotions’ to Express Better Emotions,” Age of Awareness, May 24, 2020. https://medium.com/age-of-awareness/how-to-use-wheel-of-emotions-to-express-better-emotions-8037255aa661}.  

{Tweets have a propensity to contain large amounts of information with minimal context as the tweet length is limited to 280 characters. This encourages individuals to express themselves using: 

\begin{enumerate}
    \item Hashtags, emoticons,  and mentions
    \item Misspelling and abbreviations
    \item Slang / colloquialisms 
\end{enumerate}

There can also be tweets, where no actual words are present, making it quite difficult to encode in a meaningful way for machine learning algorithms. }

Thus, the following steps detail the data-cleaning process using modules Numpy, Pandas, Emoji, and tqdm: 
\begin{enumerate}
    \item Converted emojis to their Unicode alias 
    \item Fetched hashtags and created new column to hold groups of hashtags
    \item Split contractions into separate words i.e. I'm \(\rightarrow\) I am 
    \item Removed newlines and trailing and starting spaces
    \item Removed re-tweet signals ("RT") 
    \item Removed @ symbol 
    \item Made tweets lower-case
\end{enumerate}

{A proportion of tweets may have been given an incorrect emotional label.} For instance, a happy tweet {could} say:
\begin{center}\emph{"I'm not sad, I'm really enjoying myself today."}\end{center}
However, because it contains the word "sad" this tweet could get placed into a negative emotion category by the Twitter API even if the tweet should be classified as happy. Therefore, using this data alone for classification would be problematic. 

In order to counteract this issue, all tweets were filtered through 2 pre-trained sentiment analysis networks to ensure that those misnomer tweets were removed from the dataset prior to training. The ability for these SA algorithms to classify into positive, negative, or neutral is not sufficient for the current project, which seeks to expand the set of emotions into 6 broad categories and also to learn the associations between a group of independent tweets. However, such pre-trained networks become extremely useful when leveraged as a pre-processing tool. They illuminate which tweets should be removed from the dataset before construction of the Tweet-MLN. For example, knowing that all tweets inside the Happy class should be positive, means that we simply have to remove any tweets classified as negative. 

Therefore, tweets were removed from the dataset if the sentiment label predicted by the networks did not match the current sentiment label of the data shown in Table 1. Additionally, tweets were removed if the SA pre-trained networks classified them as neutral, since that is intuitively more a lack of emotion, and therefore not relevant to the current body of research. 

{The two pre-trained models were BERTSent \cite{govindarajan2020help} which was trained on the SemEval 2017 corpus (39k plus tweets) and XLM-roBERTa-base model \cite{conneau2019unsupervised} trained on 198 million tweets.}
Both pre-trained models {were fine-tuned for sentiment analysis and} were variations of the Bidirectional Encoder Representations from Transformers (BERT) networks. BERT was developed by Google in 2018 \cite{devlin2018bert} and is used by many as a base model for sentiment analysis tasks since all the code is open-source \cite{devlin2018open}. BERT uses basic neural networks such as the recurrent neural networks (RNN) and convolution neural networks (CNN) but the encoder-decoder models work together in parallel increasing the speed of training. The structure of BERT is also made so that it can be used flexibly across several different NLP problems. Fine-tuning of pre-trained BERT networks have been used for a variety of different tasks including patent classification \cite{lee2020patent}, innovation detection \cite{chen2021leveraging}, and sentiment analysis of stock data \cite{sousa2019bert}. 

\begin{table}[hbt!]
\caption{Labels For Sentiment Analysis}
\label{table:labelsSA}
\centering
    \begin{tabular}{p{3cm}p{3cm}}
     \toprule
     \multicolumn{2}{c}{Sentiment Analysis Pre-processing} \\
     \midrule
     Tweet Emotion&Sentiment Label\\
     \hline
     Angry& Negative\\
     Bad& Negative\\
     Fearful& Negative\\
     Happy& Positive\\
     Sad& Negative\\
     Surprised& Positive\\
     \bottomrule
    \end{tabular}
\end{table}

In this way, the research leverages traditional sentiment analysis techniques in order to curate a dataset with minimal incorrectly labeled individual tweets, before construction of the Tweet-MLNs. 

Additionally, to test the validity of this method, we trained a Long-Term Short-Term Memory (LSTM) network on these sentiment analysis filtered tweets, and beta-tested the model in a real-life scenario, through platform release with PulseTech.io. The LSTM architecture helps to relieve some of the pitfalls with a traditional RNN namely the vanishing gradient problem and has high success when used for text-classification problems \cite{yao2019graph}. The accuracy based on beta-testing response with real people and real tweets was over 90\% providing evidence for the robustness of the dataset used for training the graph networks. 

After validating the dataset to ensure that tweets were placed into the correct emotion category, the data was shuffled within its' own class and split into sets of 300 tweets that contained the same class label. In total there were 900,000 tweets, with 150,000 tweets per class and a total of 500 Tweet-MLNs per class, bringing the total number of Tweet-MLNs to 3000. A train-test split of 80-20 was applied. Training and testing was run in Google Cloud Platform (GCP) on a Vertex AI cluster. It was trained for 100 epochs using one NVIDIA Tesla T4 x 1 GPU. The machine type used was n1-highmem-8 (8 vCPUs, 52 GB RAM).

\subsection{Baseline Ablation Study} \label{sub:ablation}
Since, to our knowledge, no other algorithms have been established for group-level prediction of emotions on a set of independent tweets, it is difficult to create a baseline from which to compare validation results of the MLTA. 

Additionally, the structure of the Tweet-MLNs is so specifically curated to social media text from Twitter. Meaning we cannot directly compare performances of models that test on regular text datasets. Furthermore, it is not possible to compare results of many high-performing text-classification graphs networks because they make predictions on single graph entities. Therefore, in order to analyse the network and further understand the MLTA, we perform an ablation study, in which we chose three different graph convolution layers to compare the evaluations of the MLTA trained on the Tweet-MLNs. 

There are 3 benefits to this ablation study. The first being that we can understand which graph convolution best suits a Tweet-MLN for the graph-level classification task. The second, being that we can analyse the strength of using a MLN and determine whether this architecture is useful for representing social media text. Third, we can establish a new baseline for graph-level classification on MLNs using a social media text dataset for emotion analysis, which can be referenced in the future. 

{Moreover, while we are unable to make direct comparisons between the predictions of the MLTA and other models, we can take the output of the MLTA on small subsets of tweets (for example groups of two tweets) and translate them back into their basic sentiment form. For example, taking an "angry" prediction from the MLTA and classifying it as "negative". Then comparing the results against state-of-the-art sentiment analysis models such as the BERT based models using during the filtering stage. In this way we can determine if the model is able to perform at the same level as current sentiment analysis algorithms that make highly accurate predictions on single pieces of text.}

{In order to control such granularity with the MLTA model (i.e we need to know specifically which two tweets go into each Tweet-MLN in order to track the correct sentiment), 200 tweets containing 32-40 tweets per class were manually extracted and pairs of tweets were constructed into small Tweet-MLNs. Then predictive performance for the coupled tweets were compared against state-of-the-art sentiment analysis models (see Table \ref{table:SAComparisions})}

\subsection{Evaluation Metrics}

In order to evaluate quality of prediction, we calculated the F1 score, which is the harmonic mean between the evaluation metrics recall and precision. {Therefore, the F1 score can be calculated using equation \ref{eq:F1}:}

{\begin{equation}\label{eq:F1}
    F1 = \frac{2 * Precision * Recall}{Precision + Recall} 
\end{equation}}

It takes the number of true positives (TP) where a prediction matches the positive label, true negatives (TN) or when the prediction is negative and the label is negative, false positives (FP), in which the prediction is positive but the label is negative and false negatives (FN) where the prediction is negative but the label is positive. Precision is the ratio of TP over total number of positive labels {as shown below in equation \ref{eq:precision}:} 

{\begin{equation}\label{eq:precision}
    Precision = \frac{TP}{TP + FP}
\end{equation}}

and recall is the ratio of TP over the sum of TP and FN {in equation \ref{eq:recall}}. 

{\begin{equation}\label{eq:recall}
   Recall = \frac{TP}{TP + FN}
\end{equation}}

\begin{figure*}[t!]
    \includegraphics[height=0.8\textheight,width=\textwidth,keepaspectratio]{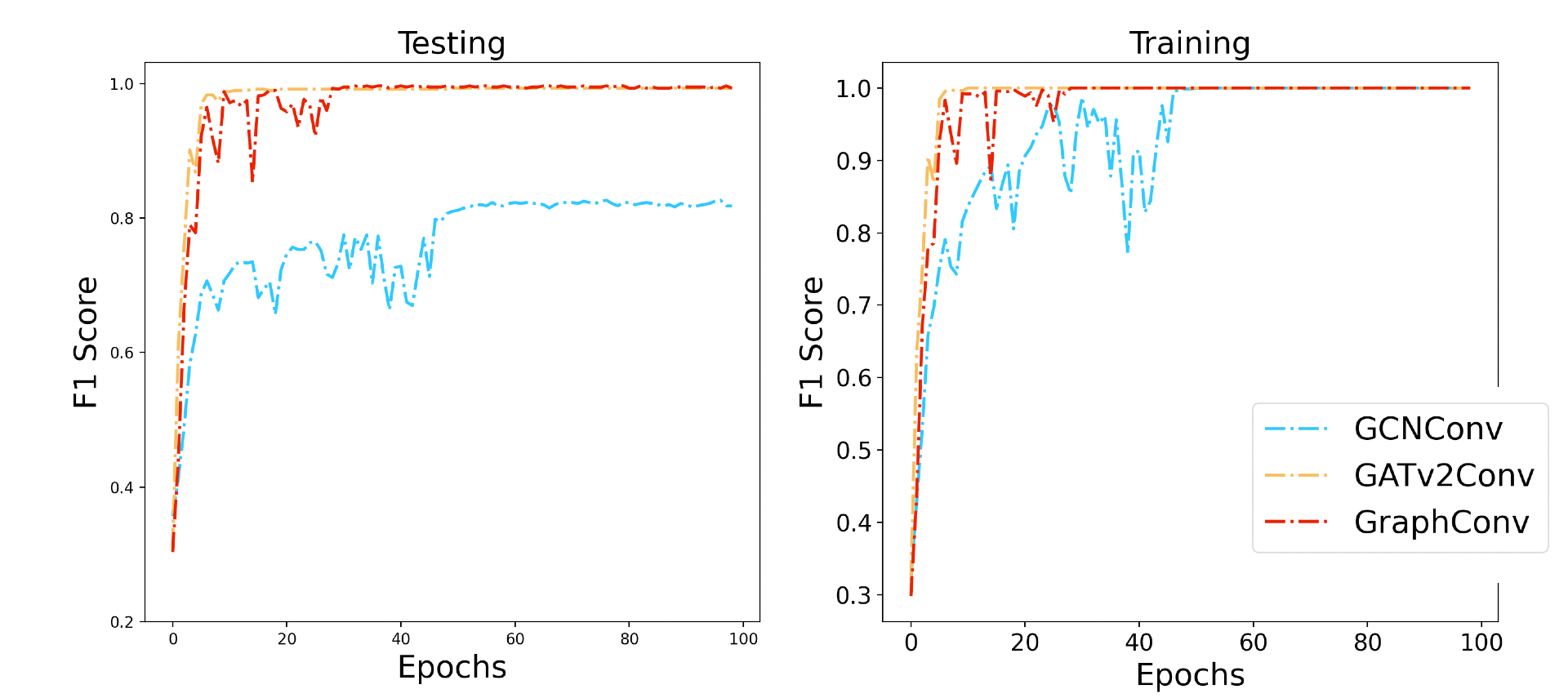}
    \caption{Testing \& Training Learning History}
    \label{fig:tt}
\end{figure*}

\section{Results}\label{sec:r}
\begin{table}[hbt!]
\caption{Comparison of Graph Convolutions for MLTA.} 
\label{table:graphComparisions}
\begin{center}
\begin{tabular}{c c c}
 \toprule
 Convolution Type & Information & F1 Test Score  \\ [0.5ex] 
 \midrule
 GCNConv & 2 Layers & 83\%   \\  [1ex]
 GATv2Conv & 2 Layers, 5 heads & 98\%  \\ [1ex]
 GraphConv & 2 Layers & 99\% \\  [1ex]
 \bottomrule
\end{tabular}
\end{center}
\end{table}

This study shows evidence that the Tweet-MLN can retain useful information specifically for the domain of sentiment analysis using social media text. Such that the combined feature information of the three layers in the MLN passed though a GNN, can discern the difference between a group of tweets that are in 6 different emotion classes. 

This can be see in Table \ref{table:graphComparisions} which demonstrates a high F1 score during evaluation despite which convolutional layers were used to train the MLTA. GraphConv was able to reach a evaluation F1 score of 99\%, and while GATv2Conv obtained a 98\% F1 score, this model was 5xs slower than GraphConv at an average of 37 minutes per epoch compared to GraphConv at an average 7 minutes per epoch. Therefore, GraphConv is the suggested model to use for prediction of future Tweet-MLNs when computational resources are an issue, without having to sacrifice performance. This discovery is useful for future implementations of a similar network architecture, because there are not many studies which evaluate the success of different graph convolutions in relation to prediction of MLNs. {Additionally, a stable learning curve over the duration of the 100 epochs shown in Figure \ref{fig:tt} suggests that the model did not overfit to the training data. While the increase in F1 score after only 20 epochs may seem quick, due to the volume of data the speed at which the model learned is not unreasonable.}

A GCN using the Chebyshev polynomial can help to speed up training time and reduce the computational costs associated. However, GCNs can have trouble capturing higher level interaction between nodes and introduces truncation errors. Therefore, it is possible that due to the complicated architecture of the Tweet-MLN, the GCNConv failed to capture and learn the full spectrum of relationships across the different layers, making it the least successful graph convolution in the ablation study. 

Additionally, the attention mechanism present in the GATv2Conv seemed to outperfrom the GCNConv. The attention mechanism was originally invented by \cite{larochelle2010learning} for computer vision with the idea that it is useful to examine different sections of the image to accumulate more information. The same principles can be applied to graph objects where we can look at all the node information, which in turn improves the learning process of the network. 

The computation of node information is done implicitly rather than explicitly as it is in the GCNConv. Therefore, using GATv2Conv we can gain a deeper understanding of each node and its' relative performance. However, while the F1 score of the GATv2Conv, was higher than the GCNConv, it was much slower compared to the GraphConv which was also able to achieve a slightly higher performance. This may be a result of the propagation process of information in the MLTA. As the information from the Tweet-MLN is not only related to its first order neighbourhood, but also to its higher order nodes from the other two layers. This makes GraphConv more advantageous for classification since higher order messages are more expressive when using GraphConv compared to other models like GCNConv and GATv2Conv. 

{\subsection{Baseline Results}}

{\begin{table}[hbt!]
\caption{Baseline Comparisons of Sentiment Analysis Against MLTA} 
\label{table:SAComparisions}
\begin{center}
\begin{tabular}{c c c}
 \toprule
 Model & Convolution & F1 Score  \\ [0.5ex] 
 \midrule
 BERTSent & - & 92\%   \\  [1ex]
 XML-ro-BERTa-base-model & - & 94\%  \\ [1ex]
 MLTA & GraphConv & 36\% \\  [1ex]
 MLTA & GCNConv & 80\% \\ [1ex] 
 MLTA & GATv2Conv & 84\% \\ [1ex]
 \bottomrule
\end{tabular}
\end{center}
\end{table}}
{The results shown in Table \ref{table:SAComparisions} of the baseline comparison against state-of-the-art sentiment analysis models reveal some interesting characteristics of the different graph convolution layers. Since the MLTA was designed for prediction on sets of 300 tweets, it makes sense that not all models were able to perform optimally after a reduction in graph size. The GraphConv only achieved an F1 score of 36\%, in comparison to the other two graph convolution layers which both received F1 scores of at least 80\% which is closer to the state-of-the-art sentiment analysis models. It seems that the GraphConv in particular does not adapt as well as the GCNConv and GATv2Conv when there are drastic changes to the number of nodes in the MLTA. While the weight placed on the higher order messages may help to increase the model performance on graphs of a similar and larger structure, it may simultaneously hinder the model's capability to be flexible towards smaller sized graphs as the benefit of higher-node information is lost.}

{Therefore, an additionally suggestion for users of the MLTA, is to use the GraphConv given data inputs of a similar size in production, and the GCNConv if the user plans on having a set of tweets with an unknown size as their input.}

{However, the most positive and most relevant finding is that, even with the detriment of reducing the graph size drastically, using a version of the MLTA on smaller subsets of tweets, we are able to achieve high levels of predictive accuracy when determining if a couplet of tweets is either positive or negative. This elucidates the power of the MLTA, as it not only is able to extract group level information, however, on a smaller scale, it can perform the same tasks as other well-supported sentiment analysis models.}

\section{Conclusion}\label{sec:c}
In summation, the MLTA, a novel approach to group-level emotion analysis of Twitter data was developed and was able to achieve high-levels of predictive performance during testing.  

{The baseline study conducted in this paper suggests that versions of the MLTA can perform well at traditional sentiment analysis tasks compared to state-of-the-art sentiment analysis models.} However, the MLTA can be seen as an extended branch of SA as it is able to not only classify the overall emotion of a group of independent texts, however it also goes beyond traditional SA categories by predicting six different emotions: Angry, Bad, Fearful, Sad, Happy, and Surprised. Moreover, this paper highlights the strength of using a multi-layered graph representation of social media text. It introduces a Tweet-MLN that is able to capture the nuances of Twitter data and allow for accurate prediction of group-level emotions. Such a network will be useful for future researchers who are looking to model complex relationships such as social media text using graph structures.

An interesting route for future research would be to investigate how group-level classification of independent texts evolve over time. Additionally, looking towards different types of MLNs such as the multi-plex MLN which would have allowed for the three layers to be connected by edges from other layers. This is an interesting development as that means, it would be possible to map the hashtag layer directly back to the text layer. It is important to contribute new studies to field of MLN research as there is lacking use of this complex network, despite its' flexibility and power.

\subsection{Acknowledgements}
We would like to acknowledge the contributions of PulseTech.io which fostered the environment to conduct this research and provided funding. 
\bibliographystyle{IEEEtran}
\bibliography{IEEEabrv,references}

\begin{IEEEbiography}[{\includegraphics[width=1in,height=1.25in,clip,keepaspectratio]{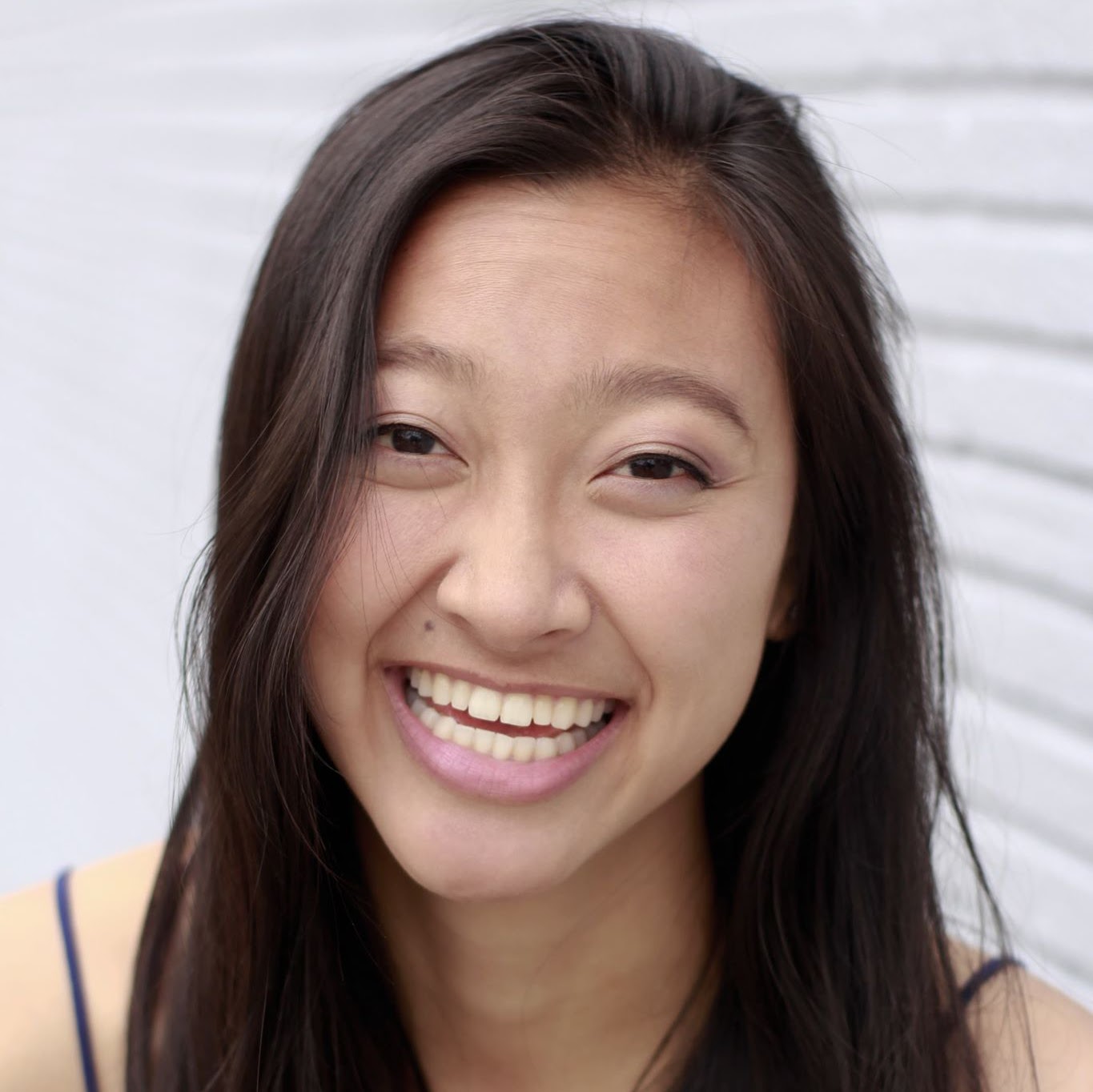}}]{Anna Nguyen} is currently working in industry as a Data Scientist. She has an MSc in Data Science from Birkbeck University and an MSc in Psychological Sciences from University College London (UCL). Professional portfolio includes leading projects that utilize natural language processing (NLP), graph modeling, and back-end development using Django. Previous research focusses on Generative Adversarial Networks (GANs) and expansion of use to commercial industries. 
\end{IEEEbiography}

\begin{IEEEbiography}[{\includegraphics[width=1in,height=1.25in,clip,keepaspectratio]{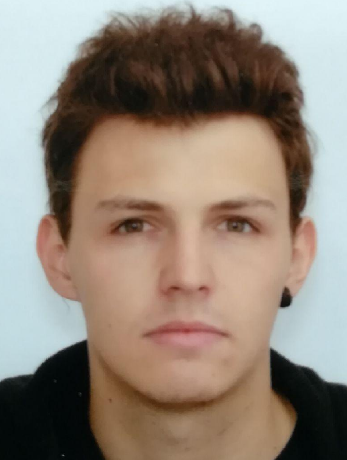}}]{Antonio Longa} is a PhD student at the University of Trento working in the Mobile and Social Computing Laboratory, Fondazione Bruno Kessler in Trento (IT) and in the Structured Machine Learning (SML) Group at the University of Trento. He is currently a visiting PhD student at the University of Cambridge (UK). My research activity focuses on Complex Systems, Temporal Networks and explainable artificial intelligence.
\end{IEEEbiography}

\begin{IEEEbiography}[{\includegraphics[width=1in,height=1.25in,clip,keepaspectratio]{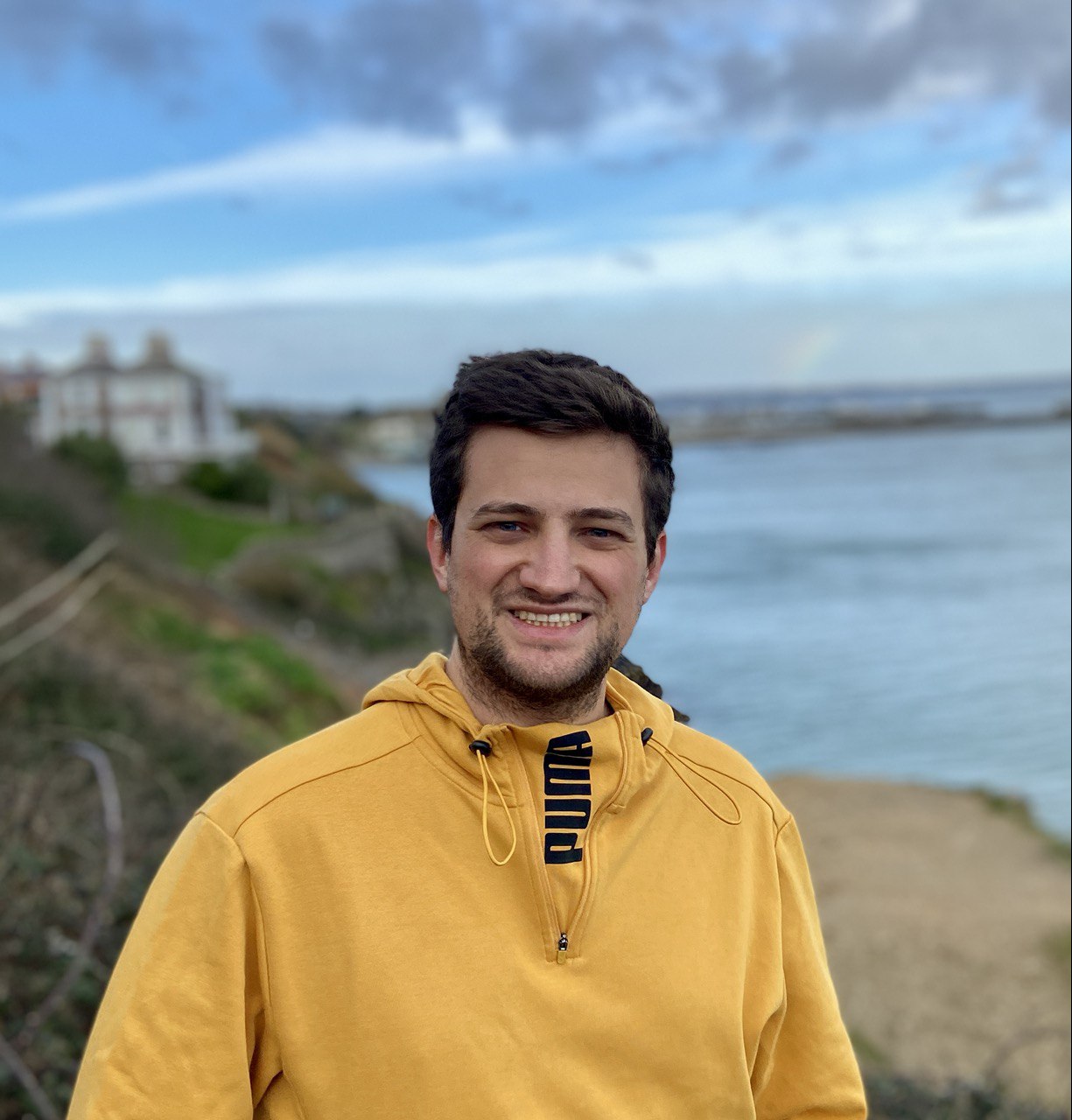}}]{Massimiliano Luca} is a Ph.D. student in the Faculty of Computer Science at the Free University of Bolzano and in the Mobile and Social Computing Lab at Fondazione Bruno Kessler. He is also an AI consultant at Pulse.io.  He is broadly interested in computational sustainability and machine learning methods to predict and generate human mobility. Previously, he was a research intern at Centro Ricerche Fiat (now Stellantis) and Telefonica Research.
\end{IEEEbiography}

\begin{IEEEbiography}[{\includegraphics[width=1in,height=1.25in,clip,keepaspectratio]{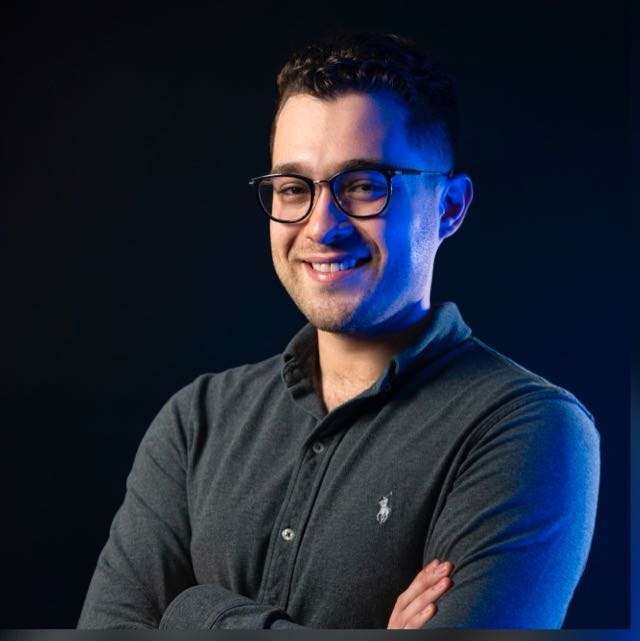}}]{Joe Kaul} is a marketing professional with almost a decade of experience in senior management in the industry. Kaul founded Pulse in 2020 with the intention of providing clearer insights into emotional intent of consumers by leveraging big data insights. Kaul has scaled his marketing agency group to a multi-million revenue group, with specific expertise in scaling return on investment by enhancing customer journey and conveying messaging in a compelling, emotionally evocative way.
\end{IEEEbiography}

\begin{IEEEbiography}[{\includegraphics[width=1in,height=1.25in,clip,keepaspectratio]{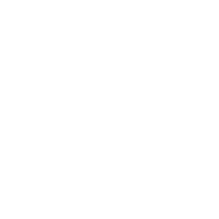}}]{Gabriel Lopez} is a self-taught Computer Scientist working at Pulse.io. He has worked in the past in Senior and Managerial positions in Machine Learning. He is broadly interested in all aspects of predicting human behavior and complex systems. His research activities involve developing new systems to reduce the unpredictability of the non-deterministic systems using neural networks.
\end{IEEEbiography}

\EOD
\end{document}